%% file: main.tex
\documentclass{article}

\usepackage{techreport}
\usepackage{geometry}
\usepackage{multirow}
\usepackage{graphicx}
\usepackage{multicol}
\usepackage{hyperref}
\usepackage{adjustbox}
\usepackage{caption}
\usepackage{subcaption}
\usepackage{booktabs}
\usepackage{amsmath}

\usepackage[utf8]{inputenc} 
\usepackage[T1]{fontenc}    
\usepackage{hyperref}       
\usepackage{url}            
\usepackage{booktabs}       
\usepackage{amsfonts}       
\usepackage{nicefrac}       
\usepackage{microtype}      
\usepackage{xcolor}         
\usepackage{threeparttable} 

\usepackage{tikz}           

\usepackage{makecell}
\usepackage[table]{xcolor}

\definecolor{lightblue}{RGB}{234,244,255}
\definecolor{darkgreen}{RGB}{0,135,0}
\definecolor{darkred}{RGB}{180,30,30}

\title{MolParser-Mobile: Ultrafast OCSR System for Large-Scale Chemical Literature Mining}

\author{
    Xi Fang \quad Haocheng Lu \quad Han Lyu \quad Chengxiang Luo \quad Linfeng Zhang \quad Guolin Ke\\  
DP Technology \\  
    {\tt\small \{fangxi, luhaocheng, lvhan, luochengxiang, zhanglf, kegl\}@dp.tech}  
}

\begin{document}

\maketitle

\vspace{-2.0em} 

\begin{center}
    \renewcommand{\arraystretch}{0.8}
    \begin{tabular}{c}
        \href{https://ocsr.dp.tech/}{\raisebox{-0.6ex}{\includegraphics[height=1.2em]{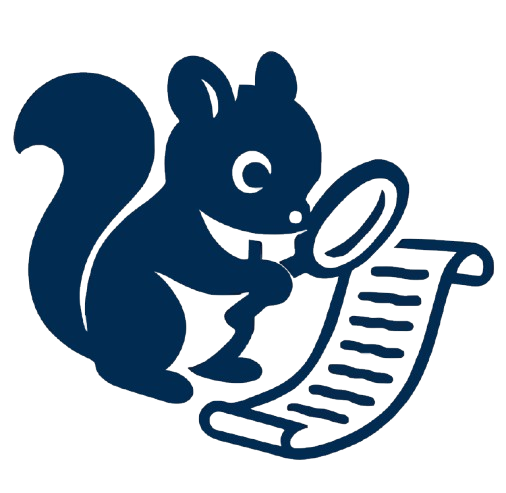}}\;\texttt{ocsr.dp.tech}} \\[0.2em]
        
        
        \href{https://github.com/dptech-corp/MolParser}{\raisebox{-0.5ex}{\includegraphics[height=1.0em]{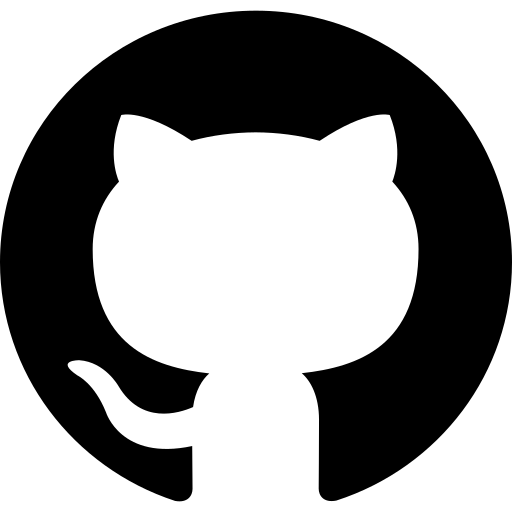}}\;\texttt{github.com/dptech-corp/MolParser}} \\[0.2em]
        
        \href{https://huggingface.co/UniParser/MolParser-Mobile}{\raisebox{-0.5ex}{\includegraphics[height=1.0em]{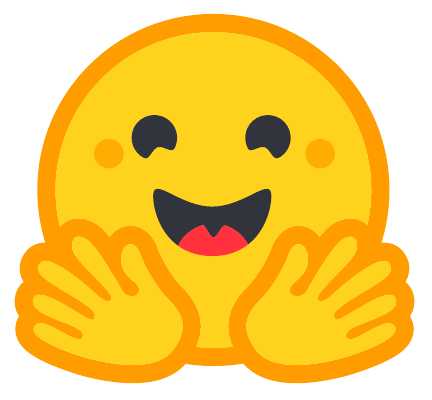}}\;\texttt{huggingface.co/UniParser/MolParser-Mobile}}
        
    \end{tabular}
\end{center}


\input{sec/0_abstract}

\input{sec/1_introduction}

\input{sec/2_moldet}

\input{sec/3_molgallery}

\input{sec/4_model}

\input{sec/5_discussion}

{
\small
\bibliography{main}
}




\end{document}

%% file: sec/0_abstract.tex
\begin{abstract}
Optical Chemical Structure Recognition (OCSR) is a fundamental component of chemical literature mining, enabling molecular database construction, reaction extraction, and AI-driven scientific discovery. Despite substantial progress in recognition accuracy with recent deep learning-based methods, inference throughput remains a critical bottleneck that limits web-scale deployment.
To address this challenge, we propose \textbf{MolParser-Mobile}, an AutoML-optimized lightweight end-to-end OCSR framework. MolParser-Mobile contains only \textbf{9.98M parameters}, while reaching a throughput of \textbf{1,520 molecules per second} on a single NVIDIA RTX 4090D GPU. Despite its compact design, it maintains competitive and, on several benchmarks, superior recognition accuracy.
\end{abstract}

%% file: sec/1_introduction.tex
\begin{table}[h]
\centering
\small
\setlength{\tabcolsep}{7pt} 

\begin{tabular}{lcccc}
\toprule
\textbf{Model} &
\textbf{Params} &
\textbf{Throughput} &
\textbf{UniParserBench} &
\textbf{BioVista} \\
\midrule

Gemma4-31B      & 31B  & OOM   & 0.073 & 0.118 \\
Qwen3.5-397B    & 397B & OOM   & 0.272 & 0.347 \\
\midrule
MolParser       & 216M & 39.8  & 0.800 & 0.703 \\

\rowcolor{lightblue}
\textbf{MolParser-Mobile}
& \hphantom{\,\textsuperscript{$\downarrow$95.4\%}}\textbf{9.98M}\,\textsuperscript{\textcolor{darkgreen}{$\downarrow$95.4\%}}
& \hphantom{\,\textsuperscript{$\uparrow$38$\times$}}\textbf{1,520}\,\textsuperscript{\textcolor{darkgreen}{$\uparrow$38$\times$}}
& \hphantom{\,\textsuperscript{$\uparrow$0.023}}\textbf{0.823}\,\textsuperscript{\textcolor{darkgreen}{$\uparrow$0.023}}
& \hphantom{\,\textsuperscript{$\uparrow$0.098}}\textbf{0.801}\,\textsuperscript{\textcolor{darkgreen}{$\uparrow$0.098}}\\

\bottomrule
\end{tabular}

\vspace{2mm}

\raggedright
\footnotesize
\emph{Throughput is measured on a single NVIDIA RTX~4090D 24GB GPU.}

\end{table}

\section{Introduction}

Recent advances in AI scientists, drug discovery, reaction prediction, molecular foundation models, knowledge graph construction, and patent mining have created an unprecedented demand for structured molecular data extracted from scientific literature. As one of the fundamental technologies for converting molecular structure images into machine-readable representations, Optical Chemical Structure Recognition (OCSR) has become a critical component of large-scale chemical knowledge mining~\cite{fang2025uni, yan2025biominer}. With the rapid growth of open-access publications and patents, modern applications routinely require parsing millions or even billions of molecular images.

Recent advances in OCSR, exemplified by atom-bond–based systems~\cite{morin2023molgrapher, qian2023molscribe} and end-to-end approaches~\cite{rajan2023decimer, fang2024molparser, fang2025uni}, have substantially advanced the accuracy of molecular image recognition. Despite these advances, existing methods remain fundamentally constrained by inference efficiency. Most prior studies have focused on enhancing model expressiveness and recognition accuracy, leaving the scalability challenge largely unresolved. Traditional OCSR systems typically require seconds per molecule, while even highly optimized end-to-end approaches such as MolParser achieve throughput only at the scale of tens of molecules per second. As a result, OCSR has become a major computational bottleneck in large-scale chemical knowledge extraction. For example, recognizing one billion molecular images at one molecule per second would consume more than 11,000 GPU-days, highlighting the urgent need for substantially more efficient OCSR solutions capable of supporting web-scale molecular data mining.

For practical chemical literature mining, inference throughput has become as important as recognition accuracy. A deployable OCSR system must simultaneously achieve high throughput, lightweight deployment, competitive recognition performance, and robustness across diverse real-world document layouts. However, satisfying these requirements cannot be achieved by merely compressing existing architectures. Efficient OCSR requires a holistic redesign of the recognition pipeline, including accurate and efficient molecule localization, large-scale real-world training data, and architecture optimization tailored specifically for high-throughput inference.

To address these challenges, we redesign the entire OCSR pipeline and propose MolParser-Mobile, an ultrafast OCSR framework for large-scale chemical literature mining. Our framework consists of three complementary components. First, we develop \textbf{MolDetv2}, a significantly improved molecule detector together with \textbf{MolDet-Bench}, a dedicated benchmark for molecule localization in scientific documents. Second, we construct \textbf{MolGallery}, a 10-million-sample real-world molecular image dataset automatically extracted from large-scale chemical literature using a multi-model consistency-based pseudo-labeling strategy. Finally, leveraging these resources, we introduce \textbf{MolParser-Mobile}, an AutoML-optimized lightweight encoder-decoder model trained on over 18 million molecular images.

MolParser-Mobile contains only 9.98M parameters, representing a 95.4\% reduction in model size compared with MolParser 1.0, while achieving 1,520 molecules per second on a single NVIDIA RTX 4090D GPU--over 38$\times$ faster than the original model. Despite its compact architecture, MolParser-Mobile achieves competitive and, on several benchmarks, superior recognition accuracy. At this throughput, one billion molecular images can be parsed within approximately one day using only eight consumer-grade GPUs, enabling practical web-scale molecular extraction for modern chemical literature mining.

%% file: sec/2_moldet.tex
\section{Molecule Detection}

Accurately localizing molecular structures in real-world chemical documents is a prerequisite for downstream optical chemical structure recognition (OCSR). This task is particularly challenging for documents containing dense text, multi-panel figures, tables, and reaction schemes~\cite{fang2024molparser,yan2025biominer,xu2022molminer}. The original MolDet~\cite{fang2024molparser} provides an open-source suite of detectors based on YOLO11~\cite{ultralytics}. However, in practical deployment, these models remain susceptible to false detections on non-chemical graphical elements, while their compact variants exhibit noticeable accuracy degradation, limiting their applicability to high-throughput document processing.

To address these limitations, we introduce \textbf{MolDetv2}\footnote{Model weights are available at \url{https://huggingface.co/UniParser/MolDetv2}}, comprising two compact detectors for complementary operating regimes. Both models adopt the 2.6-million-parameter YOLO11n architecture:

\begin{itemize}
    \item \textbf{MolDetv2-General-n} operates at a resolution of $640 \times 640$ and targets molecular depictions across a wide range of input scales, including isolated structures, reaction schemes, tables, handwritten structures, and complete document pages. Its training set contains more than 100,000 human-annotated and synthetically generated images. The training data additionally include challenging negative examples, such as chemical formulae and ball-and-stick renderings, to reduce false detections.

    \item \textbf{MolDetv2-Doc-n} operates at a resolution of $960 \times 960$ and is specialized for complete document pages containing small molecular depictions. It is trained on more than 60,000 human-annotated PDF pages collected from patents, scientific papers, and books, together with 10,000 synthetic PDF pages. Tight, edge-aligned box annotations improve localization around the visible structure extent, while document-specific negative examples reduce false detections on equations, ball-and-stick renderings, and graphical symbols.
\end{itemize}

\begin{figure*}[t]
    \centering
    \includegraphics[width=\textwidth]{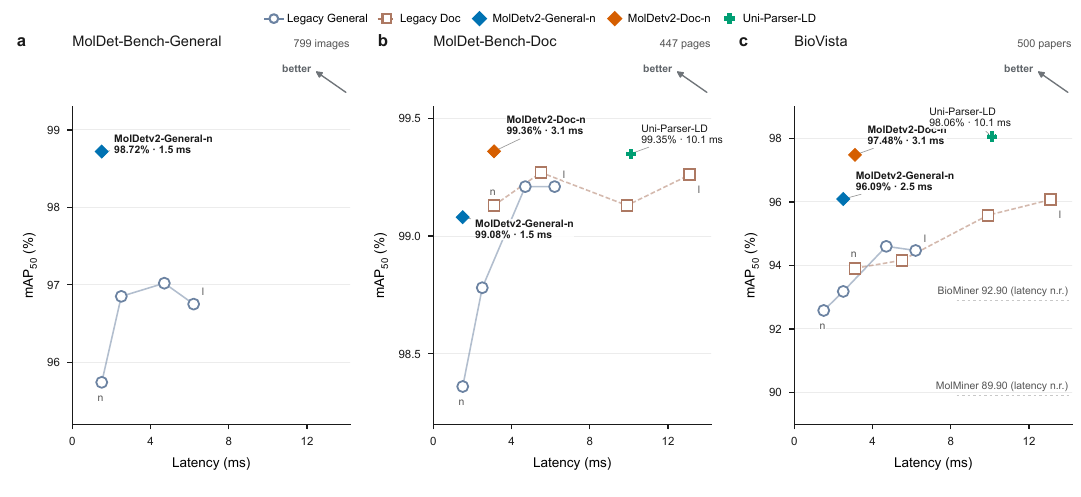}
    \caption{\textbf{Accuracy--latency trade-off of molecular structure detectors.}
    \textbf{(a)} MolDet-Bench-General evaluates molecular localization over diverse image types and scales, whereas \textbf{(b)} MolDet-Bench-Doc and \textbf{(c)} BioVista evaluate document-page inputs. MolDetv2 provides favorable Pareto operating points, achieving high $\mathrm{mAP}_{50}$ with low inference latency. Legacy MolDet variants are connected according to model scale ($n\!\rightarrow\!s\!\rightarrow\!m\!\rightarrow\!l$), while MolDetv2 and Uni-Parser-LD are shown separately. Horizontal error bars indicate latency variation, and dashed lines in \textbf{(c)} denote methods for which latency is unavailable.}
    \label{fig:moldetv2_results}
\end{figure*}

To evaluate the two operating regimes, we introduce \textbf{MolDet-Bench}\footnote{The benchmark is available at \url{https://huggingface.co/datasets/UniParser/MolDet-Bench}}, which consists of the following complementary tasks:

\begin{enumerate}
    \item \textbf{MolDet-Bench-General} evaluates multi-scale molecular localization on 799 images covering isolated and multiple structures, reaction schemes, tables, handwritten structures, and complete PDF pages. Its annotations employ adaptively expanded bounding boxes to reduce truncation when the detected regions are passed to downstream recognition models.

    \item \textbf{MolDet-Bench-Doc} evaluates document-level localization on 447 structure-bearing PDF pages containing 2,178 molecular instances. Unlike MolDet-Bench-General, this benchmark uses tight, edge-aligned bounding boxes. For additional external evaluation, we report results on BioVista~\cite{yan2025biominer}, which contains 11,212 molecular instances collected from 500 scientific papers.
\end{enumerate}

\begin{table*}[t]
    \centering
    \caption{\textbf{Representative results on MolDet-Bench.}
    We report latency-matched and strongest representative baselines; complete
    accuracy--latency comparisons are provided in Fig.~\ref{fig:moldetv2_results}.
    The best accuracy and lowest latency in each benchmark are shown in bold.
    Latency is measured in milliseconds on an NVIDIA T4 using TensorRT~10.}
    \label{tab:moldet_bench}
    \setlength{\tabcolsep}{10pt}
    \renewcommand{\arraystretch}{1.05}
    \footnotesize

    \begin{tabular}{@{}lccc@{}}
        \toprule
        Model
        & $\mathrm{mAP}_{50}\uparrow$
        & $\mathrm{mAP}_{50:95}\uparrow$
        & Latency (ms)$\downarrow$ \\
        \midrule

        \multicolumn{4}{c}{\textit{MolDet-Bench-General}} \\
        \addlinespace[2pt]
        \textbf{MolDetv2-General-n}
            & \textbf{0.9872}
            & \textbf{0.8776}
            & $\mathbf{1.5 \pm 0.0}$ \\
        MolDet-General-m
            & 0.9702
            & 0.8269
            & $4.7 \pm 0.1$ \\
        MolDet-General-l
            & 0.9675
            & 0.8349
            & $6.2 \pm 0.1$ \\
        MolDet-General-n
            & 0.9574
            & 0.8052
            & $\mathbf{1.5 \pm 0.0}$ \\

        \midrule

        \multicolumn{4}{c}{\textit{MolDet-Bench-Doc}} \\
        \addlinespace[2pt]
        \textbf{MolDetv2-Doc-n}
            & \textbf{0.9936}
            & 0.9544
            & $\mathbf{3.1 \pm 0.0}$ \\
        Uni-Parser-LD
            & 0.9935
            & \textbf{0.9679}
            & $10.1 \pm 0.2$ \\
        MolDet-Doc-s
            & 0.9927
            & 0.9531
            & $5.5 \pm 0.1$ \\
        MolDet-Doc-n
            & 0.9913
            & 0.9555
            & $\mathbf{3.1 \pm 0.0}$ \\

        \bottomrule
    \end{tabular}
\end{table*}

As shown in Table~\ref{tab:moldet_bench}, both MolDetv2 variants provide strong accuracy--latency trade-offs in their target operating regimes. On MolDet-Bench-General, MolDetv2-General-n achieves an $\mathrm{mAP}_{50}$ of 0.9872 and an $\mathrm{mAP}_{50:95}$ of 0.8776 at a latency of only $1.5\,\mathrm{ms}$. Compared with MolDet-General-n at the same latency, it improves $\mathrm{mAP}_{50}$ and $\mathrm{mAP}_{50:95}$ by 3.0 and 7.2 percentage points, respectively.

On MolDet-Bench-Doc, MolDetv2-Doc-n obtains the highest $\mathrm{mAP}_{50}$ of 0.9936, together with an $\mathrm{mAP}_{50:95}$ of 0.9544 and a latency of $3.1\,\mathrm{ms}$. It achieves accuracy comparable to the larger document-specific MolDet variants while reducing latency by approximately 44\% relative to MolDet-Doc-s. Uni-Parser-LD attains a higher $\mathrm{mAP}_{50:95}$ of 0.9679 but requires $10.1\,\mathrm{ms}$, making MolDetv2-Doc-n approximately $3.3\times$ faster. These results demonstrate that MolDetv2 offers favorable operating points for high-throughput molecular document processing.

%% file: sec/3_molgallery.tex
\section{MolGallery Dataset}

Lightweight OCSR models require large-scale and diverse training data to preserve recognition accuracy under aggressive model compression. To this end, we introduce \textbf{MolGallery}\footnote{The dataset is available at \url{https://huggingface.co/datasets/UniParser/MolGallery}}, a \textbf{10-million-scale} real-world molecular image dataset designed for training efficient OCSR systems.

Unlike existing synthetic datasets, MolGallery is constructed entirely from authentic chemical documents, including patents, scientific publications, books, and online resources. The dataset contains approximately 10 million molecular structure images (193 GB in total), capturing diverse document layouts, rendering styles, scanning artifacts, and real-world noise patterns encountered in large-scale literature mining. All samples are paired with E-SMILES annotations following the notation standard of MolParser.

Collecting manual annotations at this scale is prohibitively expensive. Therefore, we develop an automated consensus-based pseudo-labeling pipeline to construct MolGallery. Specifically, molecular images are first extracted from large-scale chemical documents using our efficient MolDetv2 detector. The extracted candidates are then recognized by three OCSR models, including two MolParser-1.0~\cite{fang2024molparser} variants and MolScribe~\cite{qian2023molscribe}.We retain only samples where all three models produce identical E-SMILES predictions, using model agreement as a strong confidence criterion.

This multi-model consensus strategy effectively removes ambiguous and low-quality samples while enabling scalable dataset construction. Although pseudo-labeling may introduce a small number of residual errors, MolGallery is primarily designed for large-scale pretraining of lightweight OCSR models. Subsequent fine-tuning on high-quality human-annotated datasets further improves recognition accuracy.

\begin{figure*}[t]
    \centering
    \includegraphics[
        width=\textwidth,
        keepaspectratio
    ]{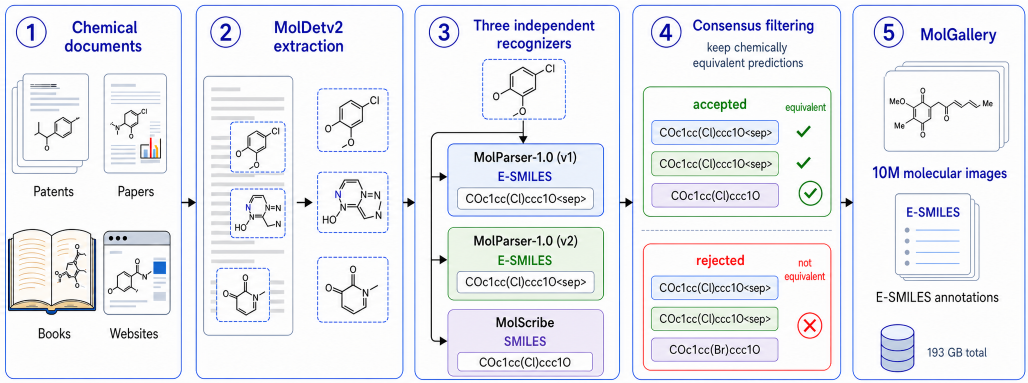}
    \caption{\textbf{MolGallery construction pipeline.} Molecular images extracted from real-world chemical documents are recognized by three OCSR models, with samples retained when all three models produce identical E-SMILES predictions.}
    \label{fig:molgallery_pipeline}
\end{figure*}

%% file: sec/4_model.tex
\section{MolParser-Mobile}
\label{sec:molparser-mobile}

Large-scale chemical literature mining needs both accurate OCSR and high
throughput.
Scaling model size is not the right path for this setting.
We introduce \textbf{MolParser-Mobile}, a 9.98M-parameter end-to-end
framework that maps a molecular diagram directly to E-SMILES.
The model does not reconstruct an atom--bond graph.
It does not rely on extra post-processing modules.
The pipeline has four stages: large-scale pre-training, supervised
fine-tuning, on-policy distillation from MolParser~1.5, and a small DPO
calibration set.

\subsection{From Molecular Diagrams to E-SMILES}
\label{sec:mobile-task}

OCSR converts a 2D chemical drawing into a machine-readable representation. While an end-to-end OCSR system can be formulated as an image-to-text generation problem, similar to image captioning, its output is fundamentally different from free-form natural language. The target is a **formal language governed by strict syntactic and chemical constraints**, where a single incorrect token can alter the molecular graph or invalidate the entire prediction.

We use E-SMILES~\cite{fang2024molparser} as the target representation:
\begin{equation}
y=\bigl[y^{\mathrm{smi}};;\texttt{<sep>};;y^{\mathrm{ext}}\bigr].
\label{eq:esmiles}
\end{equation}
Here, \(y^{\mathrm{smi}}\) is an RDKit-compatible SMILES string that serializes the molecular graph in depth-first order, while \(y^{\mathrm{ext}}\) is an optional XML-like extension for representing information that cannot be captured by standard SMILES. Specifically, it associates Markush groups, abbreviations, dummy atoms, and uncertain ring substitutions with atom or ring indices, e.g., \(\texttt{<a>}i\texttt{:R1}\texttt{</a>}\).

\subsection{Architecture}
\label{sec:mobile-arch}

MolParser~\cite{fang2024molparser} treats as decoder-only
generation.
A vision encoder emits tokens \(V=(v_1,\ldots,v_N)\in\mathbb{R}^{N\times d}\).
These tokens are prepended to the caption,
\begin{equation}
  H^{(0)}=[V;\,Y],
  \qquad
  H^{(\ell)}=\mathrm{CausalSelfAttn}\bigl(H^{(\ell-1)}\bigr).
  \label{eq:prefix}
\end{equation}
Vision and language then share one causal stream of length \(N{+}T\).

Prefix fusion mismatches the three properties above.
A causal mask is the wrong bias for a 2D graph.
Visual patches should see each other in both directions.
The encoder already provides that mixing.
As \(T\) grows, the \(N\) visual tokens recede in the context.
Softmax over \(N{+}T\) dilutes spatial evidence when the decoder later
emits \texttt{<sep>} and index tags.
Every new token also attends to \(N{+}t\) keys.
The layer cost is \(\mathcal{O}((N{+}T)^{2})\).
E-SMILES is often long, so this overhead is large.

MolParser-Mobile therefore drops the prefix.
We keep a bidirectional visual memory.
The language decoder queries it with cross-attention.
We search compact encoder--decoder configurations with AutoML, including
width, decoder depth, and training hyperparameters.
The resulting model has 9.98M parameters.

\paragraph{Vision encoder.}
OCSR inputs are predominantly line drawings rather than natural images, allowing us to use a compact vision encoder. We adopt a ResNet-V2--ViT hybrid backbone pretrained on ImageNet-21K with the AugReg recipe~\cite{steiner2021augreg}. The convolutional backbone extracts features at a spatial stride of 16, which are then projected into \(8\times8\) patches and processed by a 12-layer ViT-Tiny transformer, yielding 196 visual tokens for the autoregressive decoder.

\paragraph{Cross-attention decoder.}
The decoder is a compact BART~\cite{lewis2020bart} decoder.
It models the formal language over \(y\) only.
At step \(t\), causal self-attention reads \(y_{<t}\).
Cross-attention then looks up the image,
\begin{equation}
  \alpha_{t,i}
  =\mathrm{softmax}_{i}\!\left(
      \frac{(h_t W_Q)\,(v_i W_K)^{\top}}{\sqrt{d_h}}
    \right),
  \qquad
  c_t=\sum_{i=1}^{N}\alpha_{t,i}\,(v_i W_V).
  \label{eq:xattn}
\end{equation}
Here \(h_t\) is the decoder state.
The context \(c_t\) is fused by a residual FFN.
Visual keys and values are computed once and reused.

The design follows three principles. Self-attention is dedicated to modeling SMILES syntax, while cross-attention is dedicated to retrieving spatial information from the input image. The two interactions are kept separate rather than being mixed within a joint attention operation.

Throughout the generation of~\eqref{eq:esmiles}, the decoder queries the same visual memory \(V\). The cross-attention weights \(\alpha_{t,\cdot}\) can dynamically focus on relevant image regions as generation proceeds, enabling the model to track atoms during the depth-first traversal, revisit ring regions at closure, and associate group labels with atom or ring indices after \texttt{<sep>}.

The softmax in~\eqref{eq:xattn} is always normalized over the \(N\) visual patches, independent of the output sequence length \(T\). Thus, the attention computation per layer has complexity \(\mathcal{O}(T^2+NT)\), compared with \(\mathcal{O}((N+T)^2)\) for joint self-attention over visual and textual tokens. This distinction becomes particularly important for Markush E-SMILES, where \(T\) can be substantially larger than \(N\).

\subsection{Pre-training}
\label{sec:mobile-pt}

We first train on a large image--E-SMILES corpus.
The corpus has two sources.
\textit{MolParser-7M-Pretrain} contributes 7.72M pairs, mainly diverse
synthetic drawings~\cite{fang2024molparser}.
\textit{MolGallery} contributes 10.3M pairs from real documents.
These images are pseudo-labeled by cross-validation over multiple OCSR
models~\cite{fang2024molparser, qian2023molscribe}.
The joint set exceeds 18M pairs.

This mix matters for in-the-wild OCSR.
Synthetic data covers topology and rendering variation.
MolGallery covers document noise, crop artifacts, drawing styles, and a
higher rate of chiral and Markush cases.
The encoder learns robust visual features.
The decoder learns E-SMILES syntax at scale.

\subsection{Supervised Fine-tuning}
\label{sec:mobile-sft}

Pseudo-labels are noisy.
We therefore fine-tune on \textit{MolParser-SFT}, which contains 91.2k
human-annotated samples.
To limit catastrophic forgetting, we add a replay buffer of 200k
representative pairs from the pre-training corpus.

SFT raises precision on real literature crops.
Replay keeps coverage of rare topologies seen only in the 18M-scale
corpus.
The supervision remains teacher forcing on E-SMILES.

\subsection{On-Policy Distillation}
\label{sec:mobile-opd}

A 9.98M student still lags a larger expert on hard chiral and Markush
cases.
Uni-Parser introduces MolParser~1.5 for this regime~\cite{fang2025uni}.
MolParser~1.5 is trained on the expanded real-world corpus and is a
stronger E-SMILES expert.
We distill it into MolParser-Mobile with on-policy distillation (OPD).

The student samples trajectories \(\hat{y}\sim p_{\theta}(\cdot\mid I)\)
under its current policy.
The teacher \(p_{\phi}\) (MolParser~1.5) provides token-level
supervision on these on-policy strings.
Unlike SFT, the training distribution matches the student's own
decoding errors.
The compact model can imitate the expert on long Markush captions and
rare stereo patterns.

\subsection{Direct Preference Optimization}
\label{sec:mobile-dpo}

Pre-training labels, especially MolGallery pseudo-labels, still contain
systematic errors.
Typical failure modes include malformed Markush extensions, inconsistent
index tags, and hard chirality.
We collect 500 preference cases for targeted calibration.
Each case is a triple \((I, y^{+}, y^{-})\), where \(y^{+}\) is the
corrected E-SMILES and \(y^{-}\) is a plausible but wrong output.

We optimize Direct Preference Optimization
(DPO)~\cite{rafailov2023direct} against a frozen reference policy
\(\pi_{\mathrm{ref}}\),
\begin{equation}
  \mathcal{L}_{\mathrm{DPO}}
  =-\log\sigma\!\left(
      \beta
      \left[
        \log\frac{\pi_{\theta}(y^{+}\mid I)}{\pi_{\mathrm{ref}}(y^{+}\mid I)}
        -
        \log\frac{\pi_{\theta}(y^{-}\mid I)}{\pi_{\mathrm{ref}}(y^{-}\mid I)}
      \right]
    \right).
  \label{eq:dpo}
\end{equation}

It aligns output format on complex Markush cases and repairs residual
chiral errors after OPD.

\subsection{Efficiency and Recognition Performance}
\label{sec:mobile-eval}

We measure throughput with the native PyTorch backend.
GPU runs use FP16.
CPU runs use FP32.
No \texttt{torch.compile} is applied.
As shown in Table~\ref{tab:molparser_mobile_speed}, MolParser-Mobile
reaches 1{,}520 samples/s on an NVIDIA RTX~4090D (24~GB).
It reaches 78.11 samples/s on an Intel Xeon Platinum 8336C with 14
vCPUs.
The same 9.98M model therefore supports both GPU-scale mining and
CPU-only deployment.

\begin{table*}[t]
    \centering
    \setlength{\tabcolsep}{10pt}
    \renewcommand{\arraystretch}{1.05}
    \footnotesize

    \caption{\textbf{Inference throughput of MolParser-Mobile.}
    Throughput is measured in samples per second using the native PyTorch backend (no compile, FP16 in GPU and FP32 in CPU).}
    \label{tab:molparser_mobile_speed}

    \begin{tabular}{@{}llcc@{}}
        \toprule
        Platform
        & Hardware
        & Throughput (samples/s)$\uparrow$
        & Backend \\
        \midrule

        GPU
            & NVIDIA RTX 4090D 24GB
            & 1520
            & PyTorch \\

        CPU
            & Intel Xeon Platinum 8336C, 14 vCPU, 2.30GHz
            & 78.11
            & PyTorch \\

        \bottomrule
    \end{tabular}

    \vspace{4pt}

    \caption{\textbf{Recognition performance comparison on BioVista and Uni-Parser Bench.} All metrics are reported as absolute accuracy. The best result for each evaluation metric within each benchmark is shown in bold.}
    \label{tab:molparser_mobile_accuracy}

    \begin{tabular}{@{}lcccc@{}}
        \toprule
        Model
        & Overall$\uparrow$
        & Chiral$\uparrow$
        & Markush$\uparrow$
        & Full$\uparrow$ \\
        \midrule

        \multicolumn{5}{c}{\textit{Uni-Parser Bench}} \\
        \addlinespace[2pt]

        \textbf{MolParser-Mobile}
            & \textbf{0.823}
            & \textbf{0.732}
            & \textbf{0.746}
            & 0.933 \\

        MolParser 1.0~\cite{fang2024molparser}
            & 0.800
            & 0.676
            & 0.664
            & \textbf{0.953} \\

        MolScribe~\cite{qian2023molscribe}
            & 0.417
            & 0.274
            & 0.168
            & 0.617 \\

        \midrule

        \multicolumn{5}{c}{\textit{BioVista}} \\
        \addlinespace[2pt]

        \textbf{MolParser-Mobile}
            & \textbf{0.801}
            & \textbf{0.653}
            & \textbf{0.805}
            & \textbf{0.797} \\

        MolParser 1.0~\cite{fang2024molparser}
            & 0.703
            & 0.352
            & 0.733
            & 0.669 \\

        MolMiner~\cite{xu2022molminer}
            & 0.507
            & 0.497
            & 0.185
            & 0.774 \\

        MolScribe~\cite{qian2023molscribe}
            & 0.455
            & 0.481
            & 0.156
            & 0.703 \\

        MolNexTR~\cite{chen2024molnextr}
            & 0.401
            & 0.419
            & 0.045
            & 0.695 \\

        DECIMER~\cite{rajan2023decimer}
            & 0.298
            & 0.326
            & 0.000
            & 0.545 \\

        \bottomrule
    \end{tabular}
\end{table*}

We further evaluate MolParser-Mobile on BioVista~\cite{yan2025biominer} and Uni-Parser Bench~\cite{fang2025uni}. As summarized in Table~\ref{tab:molparser_mobile_accuracy}, MolParser-Mobile consistently outperforms MolParser~1.0 on most criteria, particularly chirality and Markush recognition, with only a minor gap in full-structure accuracy. Meanwhile, it achieves a \(38\times\) GPU throughput improvement (1,520 vs.\ 39.8 samples/s).

These results demonstrate that MolParser-Mobile achieves a favorable balance between recognition accuracy and inference efficiency. The observed performance is consistent with our design choices: cross-attention is intended to facilitate spatial information retrieval, while the large-scale corpus, supervised fine-tuning, on-policy distillation, and preference optimization are designed to improve coverage and robustness under the lightweight setting. With only 9.98M parameters, MolParser-Mobile is well suited for large-scale chemical literature mining and resource-constrained deployment.

%% file: sec/5_discussion.tex
\section{Future Works}

Future work will further advance the chemical expressiveness and image-grounding capability of MolParser toward fully end-to-end optical chemical structure recognition (OCSR) in large-scale, heterogeneous scientific and patent corpora.

\textbf{Representation Expansion.}
On the representation side, we plan to extend E-SMILES to better capture complex ring-connectivity patterns, including ring closures in fused, spiro, bridged, and macrocyclic systems. We further aim to support repeated substructures and substructure multiplicity, which are commonly used in compact chemical diagrams and Markush-style representations. In addition, we will extend E-SMILES to encode query-like or uncertain bonding patterns, such as optional or ambiguous connectivity expressed through dashed or partial bond conventions in medicinal chemistry and patent literature.

\textbf{Structure-Image Grounding.}
Beyond symbolic representation, we will explore fine-grained structure-image alignment within the end-to-end OCSR framework by grounding predicted atoms and bonds to their corresponding spatial locations in the input chemical diagram. Unlike conventional atom-bond supervised approaches, this direction does not rely on explicit structural grounding annotations. Instead, it seeks to unify visual interpretability with the scalability, efficiency, and generalization capability of end-to-end molecular recognition models.

\textbf{Scaling and Robust Pretraining.}
We will further scale MolParser using substantially larger and more diverse pretraining corpora, covering broader chemical space, depiction styles, document sources, and real-world noise distributions. This expansion is expected to improve robustness across heterogeneous scientific literature, including low-quality scans and complex patent figures.

\textbf{Chemistry-aware Optimization.}
To improve chemically valid generation, particularly for stereochemically complex molecules, we plan to introduce chirality-aware reinforcement learning objectives that explicitly reward stereochemical correctness and penalize invalid molecular configurations. This will strengthen the model’s ability to preserve chemically consistent 3D-aware constraints within 2D recognition outputs.

\textbf{Adaptive Resolution.}
In addition, we will investigate adaptive image resolution strategies that dynamically allocate higher visual resolution to large or structurally dense molecules while maintaining the efficiency advantages of the end-to-end parsing pipeline.

Overall, these directions aim to make MolParser more robust, chemically expressive, and scalable for large-scale extraction of machine-readable molecular structures from diverse scientific and patent literature.